\documentclass[journal]{IEEEtran}

\usepackage[american]{babel}

\usepackage{mathtools}
\usepackage{amsmath,amssymb}
\usepackage{multirow}
\usepackage[caption=false]{subfig}
\usepackage{bm}
\usepackage{algorithm,algorithmic}

\usepackage{graphicx}
\graphicspath{{figure/}}

\usepackage[breaklinks=true,colorlinks,bookmarks=false]{hyperref}

\begin{document}
\title{Selective Convolutional Descriptor Aggregation for Fine-Grained Image Retrieval}

\author{Xiu-Shen Wei, Jian-Hao Luo, Jianxin Wu,~\IEEEmembership{Member,~IEEE}, Zhi-Hua Zhou,~\IEEEmembership{Fellow,~IEEE}
\thanks{This work was supported in part by the National Natural Science Foundation of China under Grant No. 61422203 and No. 61333014.}
\thanks{All authors are with the National Key Laboratory for Novel Software Technology, Nanjing University, Nanjing 210023, China. J. Wu is the corresponding author. E-mail: \{weixs, luojh, wujx, zhouzh\}@lamda.nju.edu.cn.}
}

\markboth{ACCEPTED BY IEEE TIP}%
{Shell \MakeLowercase{\textit{et al.}}: Bare Demo of IEEEtran.cls for IEEE Journals}

\maketitle

\begin{abstract}
Deep convolutional neural network models pre-trained for the ImageNet classification task have been successfully adopted to tasks in other domains, such as texture description and object proposal generation, but these tasks require annotations for images in the new domain. In this paper, we focus on a novel and challenging task in the pure unsupervised setting: fine-grained image retrieval. Even with image labels, fine-grained images are difficult to classify, let alone the unsupervised retrieval task. We propose the Selective Convolutional Descriptor Aggregation (SCDA) method. SCDA firstly localizes the main object in fine-grained images, a step that discards the noisy background and keeps useful deep descriptors. The selected descriptors are then aggregated and dimensionality reduced into a short feature vector using the best practices we found. SCDA is unsupervised, using no image label or bounding box annotation. Experiments on six fine-grained datasets confirm the effectiveness of SCDA for fine-grained image retrieval. Besides, visualization of the SCDA features shows that they correspond to visual attributes (even subtle ones), which might explain SCDA's high mean average precision in fine-grained retrieval. Moreover, on general image retrieval datasets, SCDA achieves comparable retrieval results with state-of-the-art general image retrieval approaches.
\end{abstract}


\IEEEpeerreviewmaketitle

\section{Introduction}

\IEEEPARstart{A}{fter}  the breakthrough in image classification using Convolutional Neural Networks (CNN)~\cite{CNN12}, pre-trained CNN models trained for one task (e.g., recognition or detection) have also been applied to domains different from their original purposes (e.g., for describing texture~\cite{Mircea15CVPR} or finding object proposals~\cite{Amir15ICCV}). Such adaptations of pre-trained CNN models, however, still require further annotations in the new domain (e.g., image labels). In this paper, we show that for fine-grained images which contain only subtle differences among categories (e.g., varieties of dogs), pre-trained CNN models can both localize the main object and find images in the same variety. Since no supervision is used, we call this novel and challenging task \emph{fine-grained image retrieval}.

In fine-grained image classification~\cite{Ning14ECCV,Di15CVPR,Tianjun15CVPR,Marcel15ICCV,Tsung-Yu15ICCV,TIP2016weakly}, categories correspond to varieties in the same species. The categories are all similar to each other, only distinguished by slight and subtle differences. Therefore, an accurate system usually requires strong annotations, e.g., bounding boxes for object or even object parts. Such annotations are expensive and unrealistic in many real applications. In answer to this difficulty, there are attempts to categorize fine-grained images with only image-level labels, e.g.,~\cite{Tianjun15CVPR,Marcel15ICCV,Tsung-Yu15ICCV,TIP2016weakly}.

\begin{figure}[t]
 \centering
 \centering
 \subfloat[Fine-grained image retrieval. Two examples (``Mallard'' and ``Rolls-Royce Phantom Sedan 2012'') from the \emph{CUB200-2011}~\cite{WahCUB200_2011} and \emph{Cars}~\cite{cars} datasets, respectively.]  { \includegraphics[width=0.95\columnwidth]{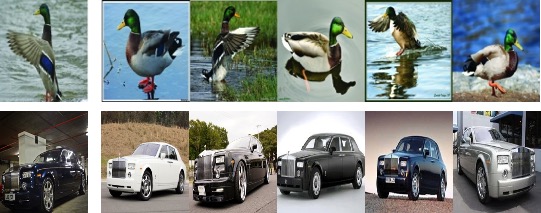} \label{fig:FGvsGN1} }
 \qquad
 \subfloat[General image retrieval. Two examples from the \emph{Oxford Building}~\cite{Oxfordbuilding} dataset.] { \includegraphics[width=0.95\columnwidth]{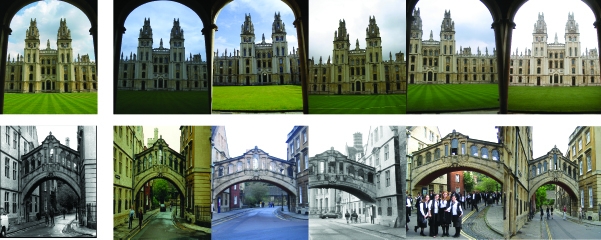} \label{fig:FGvsGN2} }
 \caption{Fine-grained image retrieval vs. general image retrieval. Fine grained image retrieval (FGIR) processes visually similar objects as the probe and gallery. For example, given an image of \emph{Mallard} (or \emph{Rolls-Royce Phantom Sedan 2012}) as the query, the FGIR system should return images of the same bird species in various poses, scales and rotations (or images of the same automobile type in various colors and angles). However, general-purpose image retrieval focuses on searching through similar images based on their similar contents, e.g., textures and shapes of the same one building. In every row, the first image is the query and the rest are retrieved images.}
 \label{fig:FGvsGN}
\end{figure}

In this paper, we handle a more challenging but more realistic task, i.e., Fine-Grained Image Retrieval (FGIR). In FGIR, given database images of the same species (e.g., birds, flowers or dogs) and a query, we should return images which are in the same variety as the query, without resorting to any other supervision signal. FGIR is useful in applications such as biological research and bio-diversity protection. As illustrated in Fig.~\ref{fig:FGvsGN}, FGIR is also different from general-purpose image retrieval. General image retrieval focuses on retrieving near-duplicate images based on similarities in their contents (e.g., textures, colors and shapes), while FGIR focuses on retrieving the images of the same types (e.g., the same species for the animals and the same model for the cars). Meanwhile, objects in fine-grained images have only subtle differences, and vary in poses, scales and rotations.

\begin{figure*}[t]
 \centering
 \includegraphics[width=0.95\textwidth]{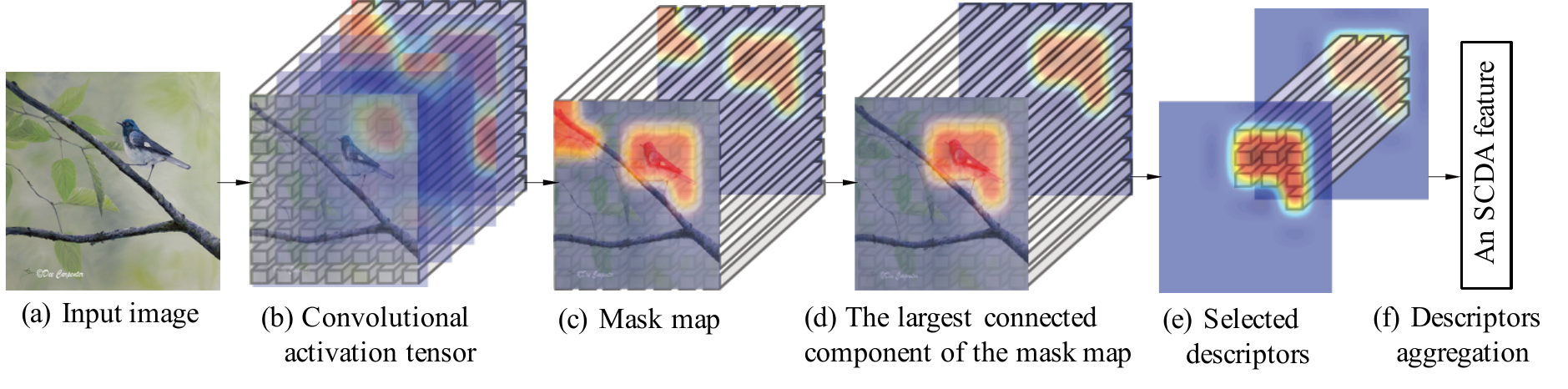}
 \caption{Pipeline of the proposed SCDA method. An input image with arbitrary resolution is fed into a pre-trained CNN model, and extracted as an order-3 convolution activation tensor. Based on the activation tensor, SCDA firstly selects the deep descriptors by locating the main object in fine-grained images unsupervisedly. Then, it pools the selected deep descriptors into the SCDA feature as the whole image representation. In the figure, (b)-(e) show the process of selecting useful deep convolutional descriptors, and the details can be found in Sec.~\ref{sec:seldes}. (This figure is best viewed in color.)}
 \label{fig:pipeline}
\end{figure*}

To meet these challenges, we propose the Selective Convolutional Descriptor Aggregation (SCDA) method, which automatically localizes the main object in fine-grained images and extracts discriminative representations for them. In SCDA, only a pre-trained CNN model (from ImageNet which is not fine-grained) is used and we use absolutely no supervision. As shown in Fig.~\ref{fig:pipeline}, the pre-trained CNN model first extracts convolution activations for an input image. We propose a novel approach to determine which part of the activations are useful (i.e., to localize the object). These useful descriptors are then aggregated and dimensionality reduced to form a vector representation using practices we propose in SCDA. Finally, a nearest neighbor search ends the FGIR process.

We conducted extensive experiments on six popular fine-grained datasets (\emph{CUB200-2011}~\cite{WahCUB200_2011}, \emph{Stanford Dogs}~\cite{Khosla11stanforddogs}, \emph{Oxford Flowers 102}~\cite{Flowers08}, \emph{Oxford-IIIT Pets}~\cite{Pets12}, \emph{Aircrafts}~\cite{airplanes} and \emph{Cars}~\cite{cars}) for image retrieval. Moreover, we also tested the proposed SCDA method on standard general-purpose retrieval datasets (\emph{INRIA Holiday}~\cite{Holiday08ECCV} and \emph{Oxford Building 5K}~\cite{Oxfordbuilding}). In addition, we report the classification accuracy of the SCDA method, which only uses the image labels. Both retrieval and classification experiments verify the effectiveness of SCDA. The key advantages and major contributions of our method are:
\begin{itemize}
 \item We propose a simple yet effective approach to localize the main object. This localization is unsupervised, without utilizing bounding boxes, image labels, object proposals, or additional learning. SCDA selects only useful deep descriptors and removes background or noise, which benefits the retrieval task.%For example, SCDA's retrieval mAP on \emph{Oxford Flowers} is 77.56\%, significantly higher than the baseline without descriptor selection (70.73\%). 
 \item With the ensemble of multiple CNN layers and the proposed dimensionality reduction practice, SCDA has shorter but more accurate representation than existing deep learning based methods (cf. Sec.~\ref{sec:experiment}). For fine-grained images, as presented in Table~\ref{table:retrieval}, SCDA achieves the best retrieval results. Furthermore, SCDA also has accurate results on general-purpose image retrieval datasets, cf. Table~\ref{table:retrievalgnr}.
 \item As shown in Fig.~\ref{fig:svd}, the compressed SCDA feature has stronger correspondence to visual attributes (even subtle ones) than the deep activations, which might explain the success of SCDA for fine-grained tasks.
\end{itemize}

Moreover, beyond the specific fine-grained image retrieval task, our proposed method could be treated as one kind of transfer learning, i.e., a model trained for one task (image classification on ImageNet) is used to solve another different task (fine-grained image retrieval). It indeed reveals the reusability of deep convolutional neural networks.

The rest of this paper is organized as follows. Sec.~\ref{sec:related} introduces the related work about general deep image retrieval and fine-grained image tasks. The details of the proposed SCDA method are presented in Sec.~\ref{sec:SCDA}. In Sec.~\ref{sec:experiment}, for fine-grained image retrieval, we compare our method with several baseline approaches and three state-of-the-art general deep image retrieval approaches. Moreover, discussion on the quality of the SCDA feature is illustrated. Sec.~\ref{sec:concl} concludes the paper.

\section{Related Work}\label{sec:related}
We will briefly review two lines of related work: deep learning approaches for image retrieval and research on fine-grained images.

\subsection{Deep Learning for Image Retrieval}
Until recently, most image retrieval approaches were based on local features (with SIFT being a typical example) and feature aggregation strategies on top of these local features. Vector of Locally Aggregated Descriptors (VLAD)~\cite{jegou2010aggregating} and Fisher Vector (FV)~\cite{sanchez2013image} are two typical feature aggregation strategies. After the success of CNN~\cite{CNN12}, image retrieval also embraced deep learning. Out-of-the-box features from pre-trained deep networks were shown to achieve state-of-the-art results in many vision related tasks, including image retrieval~\cite{Ali15CVPR}.

Some efforts (e.g.,~\cite{Gong14ECCV,Artem14ECCV,Artem15ICCV,Mattis15ICCV,Yannis16arxiv,Zheng16arxiv,rmac}) studied what deep descriptors can be used and how to use them in image retrieval, and have achieved satisfactory results. In~\cite{Gong14ECCV}, to improve the invariance of CNN activations without degrading their discriminative ability, they proposed the multi-scale orderless pooling (MOP-CNN) method. MOP-CNN firstly extracts CNN activations from the fully connected layers for local patches at multiple scale levels, and performed orderless VLAD~\cite{jegou2010aggregating} pooling of these activations at each level separately, and finally concatenated the features. After that, \cite{Artem14ECCV} has extensively evaluated the performance of such features with and without fine-tuning on related dataset. This work has shown that PCA-compressed deep features can outperform compact descriptors computed on traditional SIFT-like features. Later, \cite{Artem15ICCV} found that using sum-pooling to aggregate deep features on the last convolutional layer leads to better performance, and proposed the sum-pooled convolutional (SPoC) features. Based on that, \cite{Yannis16arxiv} applied weighting both spatially and per channel before sum-pooling to create a final aggregation. \cite{rmac} proposed a compact image representation derived from the convolutional layer activations that encodes multiple image regions without the need to re-feed multiple inputs to the network. Very recently, the authors of \cite{Zheng16arxiv} investigated several effective usages of CNN activations on both image retrieval and classification. In particular, they aggregated activations of each layer and concatenated them into the final representation, which achieved satisfactory results.

However, these approaches directly used the CNN activations/descriptors and encoded them into a single representation, without evaluating the usefulness of the obtained deep descriptors. In contrast, our proposed SCDA method can select only useful deep descriptors and remove background or noise by localizing the main object unsupervisedly. Meanwhile, we have also proposed several good practices of SCDA for retrieval tasks. In addition, the previous deep learning based image retrieval approaches were all designed for general image retrieval, which is quite different from fine-grained image retrieval. As will be shown by our experiments, state-of-the-art general image retrieval approaches do not work well for the fine-grained image retrieval task.

Additionally, several variants of image retrieval were studied in the past few years, e.g., multi-label image retrieval~\cite{TIP2016multilabel}, sketch-based image retrieval~\cite{TIP2016sketch} and medical CT image retrieval~\cite{TIP2015CT}. In this paper, we will focus on the novel and challenging fine-grained image retrieval task.

\subsection{Fine-Grained Image Tasks}

Fine-grained classification has been popular in the past few years, and a number of effective fine-grained recognition methods have been developed in the literature~\cite{Ning14ECCV,Di15CVPR,Tianjun15CVPR,Marcel15ICCV,Tsung-Yu15ICCV,TIP2016weakly}.

We can roughly categorize these methods into three groups. The first group, {e.g.},~\cite{Max15NIPS, Tsung-Yu15ICCV}, attempted to learn a more discriminative feature representation by developing powerful deep models for classifying fine-grained images. The second group aligned the objects in fine-grained images to eliminate pose variations and the influence of camera position, {e.g.},~\cite{Di15CVPR}. The last group focused on part-based representations. However, because it is not realistic to obtain strong annotations (object bounding boxes and/or part annotations) for a large number of images, more algorithms attempted to classify fine-grained images using only image-level labels, e.g.,~\cite{Tianjun15CVPR,Marcel15ICCV,Tsung-Yu15ICCV,TIP2016weakly}.

All the previous fine-grained \textit{classification} methods needed image-level labels (others even needed part annotations) to train their deep networks. Few works have touched \textit{unsupervised retrieval} of fine-grained images. Wang et al.~\cite{Deepranking} proposed Deep Ranking to learn similarity between fine-grained images. However, it requires image-level labels to build a set of triplets, which is not unsupervised and cannot scale well for large scale image retrieval tasks.

One related research to FGIR is~\cite{Lingxi15TMM}. The authors of~\cite{Lingxi15TMM} proposed the fine-grained image \emph{search} problem. \cite{Lingxi15TMM} used the bag-of-word model with SIFT features, while we use pre-trained CNN models. Beyond this difference, a more important difference is how the database is constructed. 

\cite{Lingxi15TMM} constructed a hierarchical database by merging several existing image retrieval datasets, including fine-grained datasets (e.g., \emph{CUB200-2011} and \emph{Stanford Dogs}) and general image retrieval datasets (e.g., \emph{Oxford Buildings} and \emph{Paris}). Given a query, \cite{Lingxi15TMM} first determines its meta class, and then does a fine-grained image search if the query belongs to the fine-grained meta category. In FGIR, the database contains images of one single species, which is more suitable in fine-grained applications. For example, a bird protection project may not want to find dog images given a bird query. To our best knowledge, this is \emph{the first attempt to fine-grained image retrieval using deep learning.}

\section{Selective Convolutional Descriptor Aggregation}\label{sec:SCDA}
In this section, we propose the Selective Convolutional Descriptor Aggregation (SCDA) method. Firstly, we will introduce the notations used in this paper. Then, we present the descriptor selection process, and finally, the feature aggregation details will be described.

\subsection{Preliminary}
The following notations are used in the rest of this paper. The term ``feature map'' indicates the convolution results of one channel; the term ``activations'' indicates feature maps of all channels in a convolution layer; and the term ``descriptor'' indicates the $d$-dimensional component vector of activations. ``$\text{pool}_5$'' refers to the activations of the max-pooled last convolution layer, and ``$\text{fc}_8$'' refers to the activations of the last fully connected layer.

Given an input image $I$ of size $H\times W$, the activations of a convolution layer are formulated as an order-3 tensor $T$ with $h\times w\times d$ elements, which include a set of 2-D feature maps $S=\left \{S_n\right \} \left(n=1,\ldots, d\right)$. $S_n$ of size $h\times w$ is the $n$-th feature map of the corresponding channel (the $n$-th channel). From another point of view, $T$ can be also considered as having $h\times w$ cells and each cell contains one $d$-dimensional deep descriptor. We denote the deep descriptors as $X=\left \{\bm{x}_{\left(i,j\right)}\right \}$, where $\left(i,j \right)$ is a particular cell ($i\in \left \{1,\ldots,h\right \}, j\in \left \{1,\ldots,w \right \}, \bm{x}_{\left(i,j\right)}\in \mathcal{R}^{d}$). For instance, by employing the popular pre-trained VGG-16 model~\cite{vgg16} to extract deep descriptors, we can get a $7\times 7 \times 512$ activation tensor in $\text{pool}_5$ if the input image is $224\times 224$. Thus, on one hand, for this image, we have 512 feature maps (i.e., $S_n$) of size $7\times 7$; on the other hand, 49 deep descriptors of $512$-d are also obtained.

\subsection{Selecting Convolutional Descriptors}\label{sec:select}
What distinguishes SCDA from existing deep learning-based image retrieval methods is: using only the pre-trained model, SCDA is able to find \emph{useful deep convolutional features}, which in effect \emph{localizes the main object} in the image and discards irrelevant and noisy image regions. Note that the pre-trained model is \emph{not} fine-tuned using the target fine-grained dataset. In the following, we propose our descriptor selection method, and then present quantitative and qualitative localization results.

\subsubsection{Descriptor Selection}\label{sec:seldes}
After obtaining the $\text{pool}_5$ activations, the input image $I$ is represented by an order-3 tensor $T$, which is a sparse and \emph{distributed} representation~\cite{Hinton86ACCSS,BengioPAMI}. The distributed representation argument claims that concepts are encoded by a distributed pattern of activities spread across multiple neurons~\cite{Georgopoulos86}. In deep neural networks, a distributed representation means a many-to-many relationship between two types of representations (i.e., concepts and neurons): Each concept is represented by a pattern of activity distributed over many neurons, and each neuron participates in the representation of many concepts~\cite{Hinton86ACCSS,BengioPAMI}.

\begin{figure*}[t]
 \centering
 \includegraphics[width=0.98\textwidth]{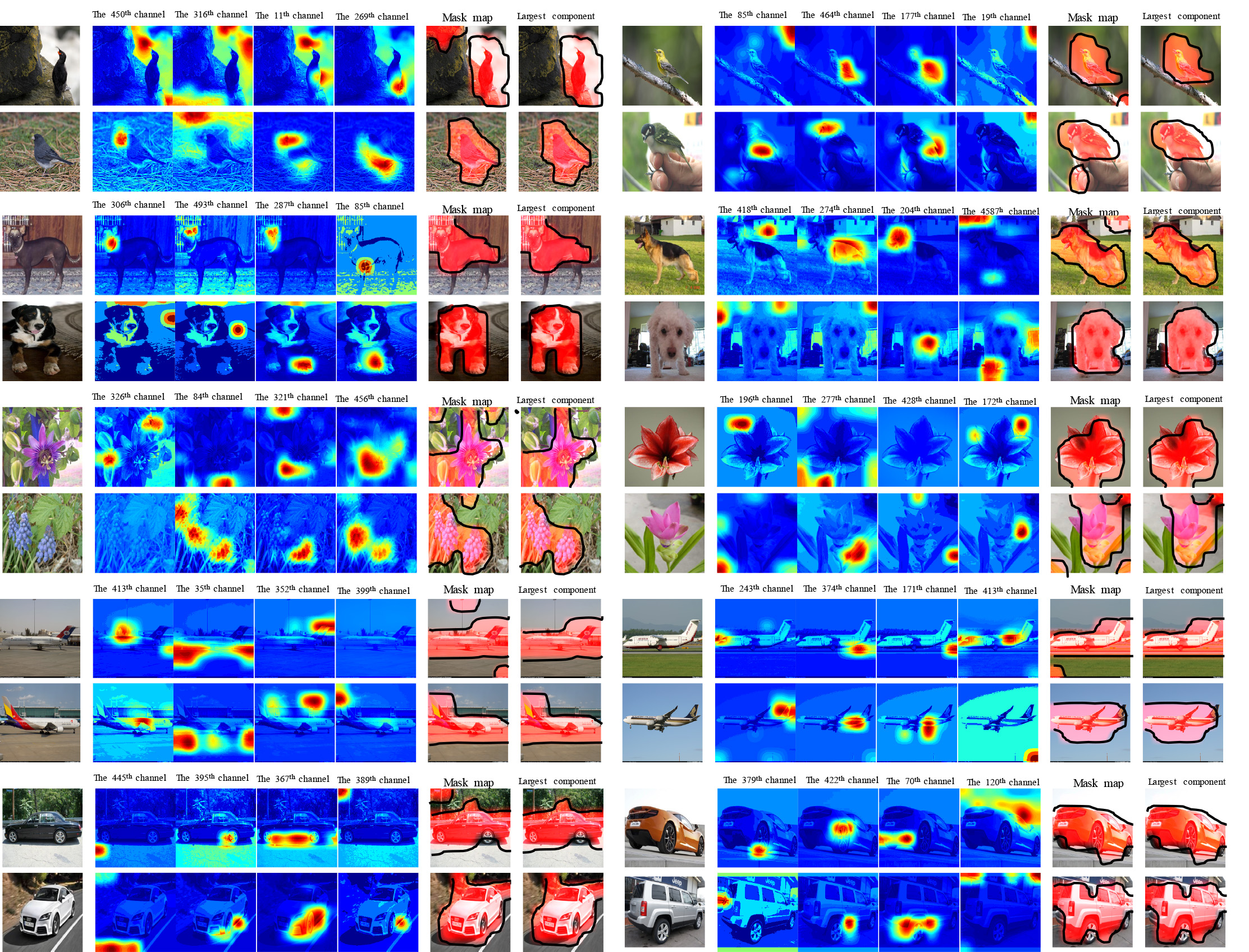}
  \caption{Sampled feature maps of fine-grained images from five fine-grained datasets (\emph{CUB200-2011}, \emph{Stanford Dogs}, \emph{Oxford Flowers}, \emph{Aircrafts} and \emph{Cars}). Although we resize the images for better visualization, our method can deal with images of any resolution. The first column of each subfigure are the input images, and the randomly sampled feature maps are the following four columns. The last two columns are the  mask maps $M$ and the corresponding largest connected component $\widetilde{M}$. The selected regions are highlighted in red with the black boundary. (The figure is best viewed in color.)}
 \label{fig:featmaps}
\end{figure*}

In Fig.~\ref{fig:featmaps}, we show some images taken from five fine-grained datasets, \emph{CUB200-2011}~\cite{WahCUB200_2011}, \emph{Stanford Dogs}~\cite{Khosla11stanforddogs}, \emph{Oxford Flowers 102}~\cite{Flowers08}, \emph{Aircrafts}~\cite{airplanes} and \emph{Cars}~\cite{cars}. We randomly sample several feature maps from the 512 feature maps in $\text{pool}_5$ and overlay them to original images for better visualization. As can be seen from Fig.~\ref{fig:featmaps}, the activated regions of the sampled feature map (highlighted in warm color) may indicate semantically meaningful parts of birds/dogs/flowers/aircrafts/cars, but can also indicate some background or noisy parts in these fine-grained images.

In addition, the semantic meanings of the activated regions are quite different even for the same channel. For example, in the 464th feature map for \emph{birds} on the right side, the activated region in the first image indicates the \emph{Pine Warbler}'s tail and the second does the \emph{Black-capped Vireo}'s head. In the 274th feature map for \emph{dogs}, the first indicates the \emph{German Shepherd}'s head, while the second even has no activated region for the \emph{Cockapoo}, except for a part of noisy background. The other examples of flowers, aircrafts and cars have the same characteristics. In addition, there are also some activated regions representing the background, e.g., the 19th feature map for \emph{Pine Warbler} and the 418th one for \emph{German Shepherd}. Fig.~\ref{fig:featmaps} conveys that \emph{not all deep descriptors are useful}, and \emph{one single channel contains at best weak semantic information} due to the distributed nature of this representation. Therefore, selecting and using only useful deep descriptors (and removing noise) is necessary. However, in order to decide which deep descriptor is useful (i.e., containing the object we want to retrieve), we cannot count on any single channel individually.

We propose a simple yet effective method (shown in Fig.~\ref{fig:pipeline}), and its quantitative and qualitative evaluation will be demonstrated in the next section. Although one single channel is not very useful, if many channels fire at the same region, we could expect this region to be an object rather than the background. Therefore, in the proposed method, we add up the obtained $\text{pool}_5$ activation tensor through the depth direction. Thus, the $h\times w\times d$ 3-D tensor becomes an $h\times w$ 2-D tensor, which we call the ``aggregation map'', i.e., $A = \sum_{n=1}^d S_n$ (where $S_n$ is the $n$-th feature map in $\text{pool}_5$). For the aggregation map $A$, there are $h\times w$ summed activation responses, corresponding to $h\times w$ positions. Based on the aforementioned observation, it is straightforward to say that the higher activation response of a particular position $\left(i,j\right)$, the more possibility of its corresponding region being part of the object. Additionally, fine-grained image retrieval is an unsupervised problem, in which we have no prior knowledge of how to deal with it. Consequently, we calculate the mean value $\bar{a}$ of all the positions in $A$ as the threshold to decide which positions localize objects: the position $\left(i,j\right)$ whose activation response is higher than $\bar{a}$ indicates the main object, e.g., birds, dogs or aircrafts, might appear in that position. A mask map $M$ of the same size as $A$ can be obtained as:
\begin{equation}
M_{i,j} = \left\{
\begin{aligned}
1 & &\text{if } A_{i,j} > \bar{a} \\
0 & & \text{otherwise}
\end{aligned}
\right.\,,
\end{equation}
where $\left(i,j\right)$ is a particular position in these $h\times w$ positions.

In Fig.~\ref{fig:featmaps}, the figures in the second last column for each fine-grained datasets show some examples of the mask maps for birds, dogs, flowers, aircrafts and cars, respectively. For these figures, we first resize the mask map $M$ using the bicubic interpolation, such that its size is the same as the input image. We then overlay the corresponding mask map (highlighted in red) onto the original images. Even though the proposed method does not train on these datasets, the main objects (e.g., birds, dogs, aircrafts or cars) can be roughly detected. However, as can be seen from these figures, there are still several small noisy parts activated on a complicated background. Fortunately, because the noisy parts are usually smaller than the main object, we employ Algorithm~\ref{algo:bwconn} to collect the largest connected component of $M$, which is denoted as $\widetilde{M}$, to get rid of the interference caused by noisy parts. In the last column, the main objects are kept by $\widetilde{M}$, while the noisy parts are discarded, e.g., the \emph{plant}, the \emph{cloud} and the \emph{grass}.

\begin{algorithm}[t]
\small
\caption{\small Finding connected components in binary images}
\label{algo:bwconn}
\begin{algorithmic}[1]{
\REQUIRE {A binary image $I$};
\STATE {Select one pixel $p$ as the starting point};
\WHILE {True}
\STATE {Use a flood-fill algorithm to label all the pixels in the connected component containing $p$};
\IF {All the pixels are labeled}
\STATE {Break};
\ENDIF
\STATE {Search for the next unlabeled pixel as $p$};
\ENDWHILE
\RETURN {Connectivity of the connected components, and their corresponding size (pixel numbers)}.
}\end{algorithmic}
\end{algorithm}

%\begin{figure*}[t]
% \centering
% \includegraphics[width=0.99\textwidth]{mask}
%  \caption{Visualization of the mask map $M$ and the corresponding largest connected component $\widetilde{M}$ of six fine-grained datasets (i.e., \emph{birds}, \emph{dogs}, \emph{flowers}, \emph{pets}, \emph{aircrafts} and \emph{cars}). The selected regions are highlighted in red. (Best viewed in color.)}
% \label{fig:mask}
%\end{figure*}

Therefore, we use $\widetilde{M}$ to select useful and meaningful deep convolutional descriptors. The descriptor $\bm{x}_{\left(i,j\right)}$ should be kept when $\widetilde{M}_{i,j}=1$, while $\widetilde{M}_{i,j}=0$ means the position $\left(i,j\right)$ might have background or noisy parts:
\begin{equation}
F = \left \{\bm{x}_{\left(i,j\right)}|\widetilde{M}_{i,j}=1 \right \} \,,
\end{equation}
where $F$ stands for the selected descriptor set, which will be aggregated into the final representation for retrieving fine-grained images. The whole convolutional descriptor selection process is illustrated in Fig.~\ref{fig:pipeline}b-\ref{fig:pipeline}e.

%\begin{figure*}[t]
% \centering
% \includegraphics[width=0.99\textwidth]{selectdesc}
%  \caption{Selecting useful deep convolutional descriptors. (This figure is best viewed in color.)}
% \label{fig:selectdesc}
%\end{figure*}

\subsubsection{Qualitative Evaluation}
In this section, we give the qualitative evaluation of the proposed descriptor selection process. Because four fine-grained datasets (i.e., \emph{CUB200-2011}, \emph{Stanford Dogs}, \emph{Aircrafts} and \emph{Cars}) supply the ground-truth bounding box for each image, it is desirable to evaluate the proposed method for object localization. However, as seen in Fig.~\ref{fig:featmaps}, the detected regions are irregularly shaped. So, the minimum rectangle bounding boxes which contain the detected regions are returned as our object localization predictions.

\begin{figure*}[t]
 \centering
 \includegraphics[width=0.95\textwidth]{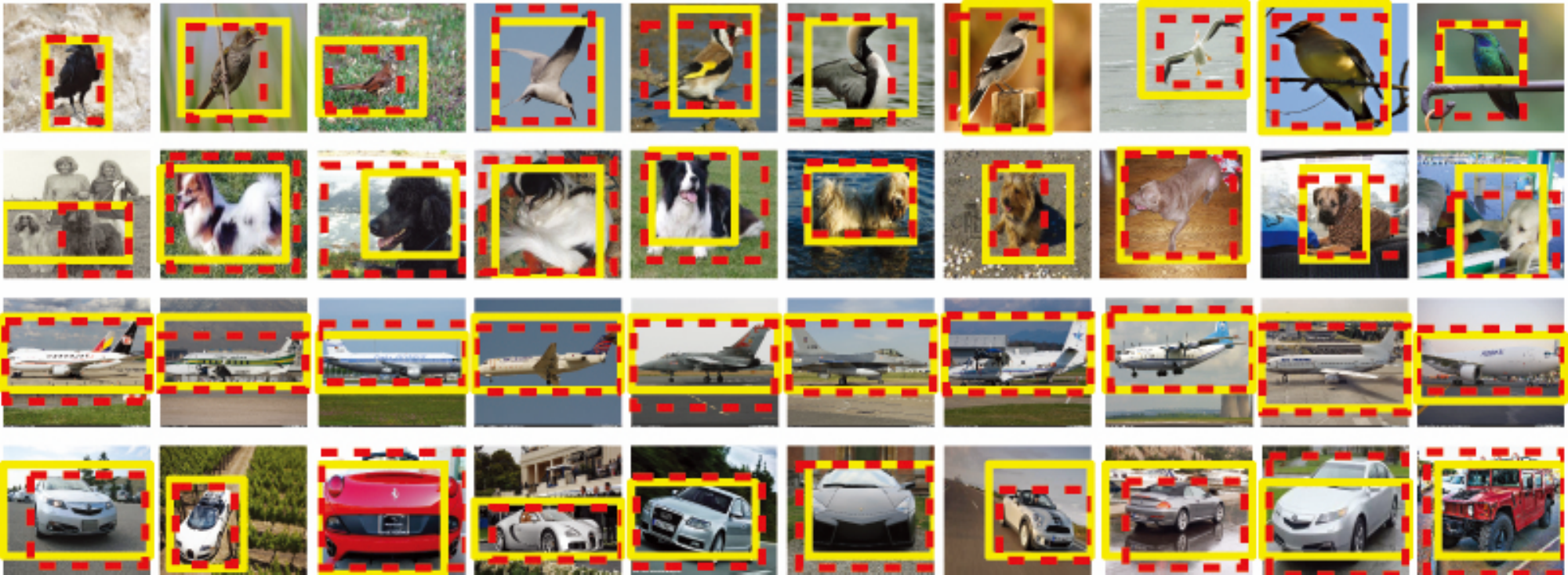}
  \caption{Random samples of predicted object localization bounding box. Each row contains ten representative object localization bounding box results for four fine-grained datasets, respectively. The ground-truth bounding box is marked as the red dashed rectangle, while the predicted one is marked in the solid yellow rectangle. (The figure is best viewed in color.)}
 \label{fig:predbbox}
\end{figure*}

We evaluate the proposed method to localize the whole-object (birds, dogs, aircrafts or cars) on their test sets. Example predictions can be seen in Fig.~\ref{fig:predbbox}. From these figures, the predicted bounding boxes approximate the ground-truth bounding boxes fairly accurately, and even some results are better than the ground truth. For instance, in the first dog image shown in Fig.~\ref{fig:predbbox}, the predicted bounding box can cover both dogs; and in the third one, the predicted box contains less background, which is beneficial to retrieval performance. Moreover, the predicted boxes of \emph{Aircrafts} and \emph{Cars} are almost identical to the ground-truth bounding boxes in many cases. However, since we utilize no supervision, some details of the fine-grained objects, e.g., birds' tails, cannot be contained accurately by the predicted bounding boxes. 

\subsubsection{Quantitative Evaluation}

We also report the results in terms of the Percentage of Correctly Localized Parts (PCP) metric for object localization in Table~\ref{table:IOU}. The reported metrics are the percentage of whole-object boxes that are correctly localized with a \textgreater$50\%$ IOU with the ground-truth bounding boxes. In this table, for \emph{CUB200-2011}, we show the PCP results of two fine-grained parts (i.e., head and torso) reported in some previous part localization based fine-grained \emph{classification} algorithms~\cite{Azizpour12ECCV,Ning14ECCV,Di15CVPR}. Here, we first compare the whole-object localization rates with that of fine-grained parts for a rough comparison. In fact, the \emph{torso} bounding box is highly similar to that of the whole-object in \emph{CUB200-2011}. By comparing the results of PCP for \emph{torso} and our \emph{whole-object}, we find that, even though our method is unsupervised, the localization performance is just slightly lower or even comparable to that of these algorithms using strong supervisions, e.g., ground-truth bounding box and parts annotations (even in the test phase). For \emph{Stanford Dogs}, our method can get 78.86\% object localization accuracy. Moreover, the results of \emph{Aircrafts} and \emph{Cars} are 94.91\% and 90.96\%, which validates the effectiveness of the proposed unsupervised object localization method.

\begin{figure}[t]
 \centering
 \subfloat[\emph{CUB200-2011}]  { \includegraphics[width=0.4\columnwidth]{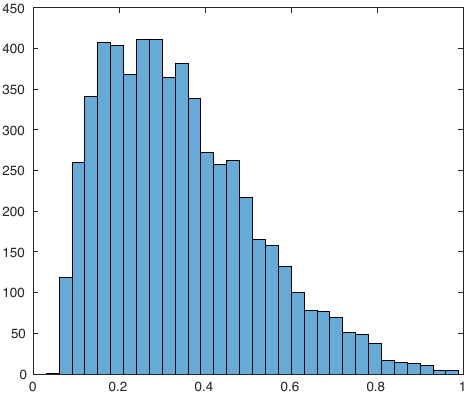} \label{fig:cub_perc} }
 \quad
 \subfloat[\emph{Stanford Dogs}] { \includegraphics[width=0.4\columnwidth]{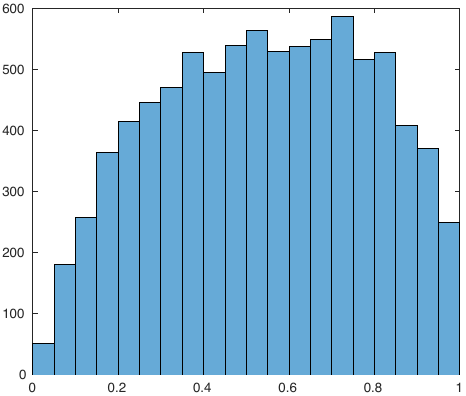} \label{fig:dog_perc} }
 \quad
 \subfloat[\emph{Aircrafts}] { \includegraphics[width=0.4\columnwidth]{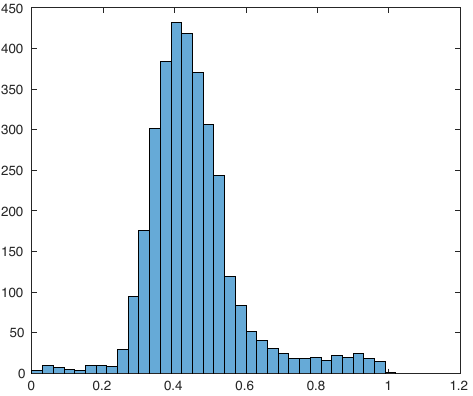} \label{fig:airplane_perc} }
 \quad
 \subfloat[\emph{Cars}] { \includegraphics[width=0.4\columnwidth]{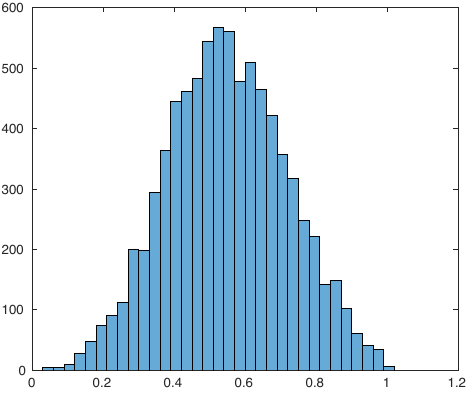} \label{fig:car_perc} }
  \caption{Percentage of the whole images covered by the ground truth bounding boxes on four fine-grained datasets. The vertical axis is the number of images, and the horizontal axis is the percentage.}
 \label{fig:percent}
\end{figure}

Additionally, in our proposed method, the largest connected component $\widetilde{M}$ of the obtained mask map is kept. We further investigate how this filtering step affects object localization performance by removing this processing. Then, based on $M$, the object localization results based on these datasets are: 45.18\%, 68.67\%, 59.83\% and 79.36\% for \emph{CUB200-2011}, \emph{Stanford Dogs}, \emph{Aircrafts} and \emph{Cars}, respectively. The localization accuracy based on $M$ is much lower than the accuracy based on $\widetilde{M}$, which proves the effectiveness of obtaining the largest connected component. Besides, we also consider these drops through the relation to the size of the ground truth bounding boxes. From this point of view, Fig.~\ref{fig:percent} shows the percentage of the whole images covered by the ground truth bounding boxes on four fine-grained datasets, respectively. It is obvious that most ground truth bounding boxes of \emph{CUB200-2011} and \emph{Aircrafts} are less than 50\% size of the whole images. Thus, for the two datasets, the drops are large. However, for \emph{Cars}, as shown in Fig.~\ref{fig:car_perc}, the percentage's distribution approaches a normal distribution. For \emph{Stanford Dogs}, a few ground truth bounding boxes cover less than 20\% image size or covering more than 80\% image size. Therefore, for these two datasets, the effect of removing the largest connected component processing could be small.

What's more, because our method does not require any supervision, a state-of-the-art unsupervised object localization method, i.e.,~\cite{Cho15CVPR}, is conducted as the baseline. \cite{Cho15CVPR} uses off-the-shelf region proposals to form a set of candidate bounding boxes for objects. Then, these regions are matched across images using a probabilistic Hough transform that evaluates the confidence for each candidate correspondence considering both appearance and spatial consistency. After that, dominant objects are discovered and localized by comparing the scores of candidate regions and selecting those that stand out over other regions containing them. As~\cite{Cho15CVPR} is not a deep learning based method, most of its localization results on these fine-grained datasets are not satisfactory, which are reported in Table~\ref{table:IOU}. Specifically, for many images of \emph{Aircrafts}, \cite{Cho15CVPR} returns the whole images as the corresponding bounding boxes predictions. While, as shown in Fig.~\ref{fig:airplane_perc}, only a small percentage of ground truth bounding boxes approach the whole images, which could explain why the unsupervised localization accuracy on \emph{Aircrafts} of~\cite{Cho15CVPR} is much worse than ours.

\begin{table*}[t]
 \caption{Comparison of object localization performance on four fine-grained datasets.} \label{table:IOU}
 \centering
 \begin{tabular}{|c|c|c|c|c|c|c|c|c|}
  \hline
  \multirow{2}{*}{Dataset} & \multirow{2}{*}{Method} & \multicolumn{2}{c|}{Train phase} & \multicolumn{2}{c|}{Test phase} & \multirow{2}{*}{Head} & \multirow{2}{*}{Torso} & \multirow{2}{*}{Whole-object} \\
  \cline{3-6} & & BBox & Parts & BBox & Parts & & & \\
  \hline
  \hline
  \multirow{5}{*}{\emph{CUB200-2011}} & Strong DPM~\cite{Azizpour12ECCV} & $\checkmark$ & $\checkmark$ & $\checkmark$ & & 43.49 & 75.15 & -- \\
  & Part-based R-CNN with BBox~\cite{Ning14ECCV} & $\checkmark$ & $\checkmark$ & $\checkmark$ & & 68.19 & 79.82 & -- \\
  & Deep LAC~\cite{Di15CVPR} & $\checkmark$ & $\checkmark$ & $\checkmark$ & & 74.00 & 96.00 & --\\
  & Part-based R-CNN~\cite{Ning14ECCV} & $\checkmark$ & $\checkmark$ & & & 61.42 & 70.68 & -- \\  
  & Unsupervised object discovery~\cite{Cho15CVPR} & & & & & -- & -- & {69.37} \\
  & {Ours} & & & & & -- & -- & {76.79} \\
\hline
\hline
 \multirow{2}{*}{\emph{Stanford Dogs}} & Unsupervised object discovery~\cite{Cho15CVPR} & & & & & -- & -- & {36.23} \\
  & {Ours} & & & & & -- & -- & {78.86} \\
\hline
\hline
 \multirow{2}{*}{\emph{Aircrafts}} & Unsupervised object discovery~\cite{Cho15CVPR} & & & & & -- & -- & {42.11} \\
  & {Ours} & & & & & -- & -- & {94.91} \\
\hline
\hline
  \multirow{2}{*}{\emph{Cars}} & Unsupervised object discovery~\cite{Cho15CVPR} & & & & & -- & -- & {93.05} \\
  & {Ours} & & & & & -- & -- & {90.96} \\
\hline
 \end{tabular}
\end{table*}

\subsection{Aggregating Convolutional Descriptors}
After the selection process, the selected descriptor set $F = \left \{\bm{x}_{\left(i,j\right)}|\widetilde{M}_{i,j}=1 \right \}$ is obtained. In the following, we compare several encoding or pooling approaches to aggregate these convolutional features, and then give our proposal.

\begin{itemize}
\item \textbf{Vector of Locally Aggregated Descriptors} (VLAD)~\cite{jegou2010aggregating} is a popular encoding approach in computer vision. VLAD uses $k$-means to find a codebook of $K$ centroids $\left\{\bm{c}_1,\ldots,\bm{c}_K \right\}$ and maps $\bm{x}_{\left(i,j\right)}$ into a single vector $\bm{v}_{\left(i,j\right)} = \left[\bm{0}~\ldots~\bm{0}~ \bm{x}_{\left(i,j\right)}-\bm{c}_k~\ldots~\bm{0}  \right]\in \mathcal{R}^{K\times d}$, where $\bm{c}_k$ is the closest centroid to $\bm{x}_{\left(i,j\right)}$. The final representation of $F$ is $\sum_{i,j} \bm{v}_{\left(i,j\right)}$.

\item \textbf{Fisher Vector} (FV)~\cite{sanchez2013image}. The encoding process of FV is similar to VLAD. But it uses a soft assignment (i.e., Gaussian Mixture Model) instead of using $k$-means for pre-computing the codebook. Moreover, FV also includes second-order statistics.

\item \textbf{Pooling approaches}. We also try two traditional pooling approaches, i.e., global average-pooling and max-pooling, to aggregate the deep descriptors, i.e.,
\begin{align}
 \nonumber
 \bm{p}_{\rm avg} &= \frac{1}{N} \sum\nolimits_{i,j} \bm{x}_{\left(i,j\right)} \,,  \\
 \nonumber
 \bm{p}_{\rm max} &= \max_{i,j} \bm{x}_{\left(i,j\right)} \,,
\end{align}
where $\bm{p}_{\rm avg}$ and $\bm{p}_{\rm max}$ are both $1\times d$ dimensional. $N$ is the number of the selected descriptors.
\end{itemize}

After encoding or pooling the selected descriptor set $F$ into a single vector, for VLAD and FV, the square root normalization and $\ell_2$-normalization are followed; for max- and average-pooling methods, we do $\ell_2$-normalization (the square root normalization did not work well). Finally, the cosine similarity is used for nearest neighbor search. We use two datasets to demonstrate which type of aggregation method is optimal for fine-grained image retrieval. The original training and testing splits provided in the datasets are used. Each image in the testing set is treated as a query, and the training images are regarded as the gallery. The top-$k$ mAP retrieval performance is reported in Table~\ref{table:poolcompa}.

For the parameter choice of VLAD/FV, we follow the suggestions reported in~\cite{Wei15ICCVW}. The number of clusters in VLAD and the number of Gaussian components in FV are both set to 2. As shown in the table, larger values lead to lower accuracy. Moreover, we find the simpler aggregation methods such as global max- and average-pooling achieve better retrieval performance comparing with the high-dimensional encoding approaches. These observations are also consistent with the findings in~\cite{Artem15ICCV} for general image retrieval. The reason why VLAD and FV do not work well in this case is related to the rather small number of deep descriptors that need to be aggregated. The average number of deep descriptors selected per image for \emph{CUB200-2011} and \emph{Stanford Dogs} is 40.12 and 46.74, respectively. Then, we propose to concatenate the global max-pooling and average-pooling representations, ``avg$\&$maxPool'', as our aggregation scheme. Its performance is significantly and consistently higher than the others. We use the ``avg$\&$maxPool'' aggregation as ``\emph{SCDA feature}'' to represent the whole fine-grained image.

\begin{table}[t]
 \caption{Comparison of different encoding or pooling approaches for FGIR. The best result of each column is marked in bold.} \label{table:poolcompa}
 \centering
 \setlength{\tabcolsep}{5pt}
 \begin{tabular}{|c|c|c|c|c|c|}
  \hline
  \multirow{2}{*}{Approach} & \multirow{2}{*}{Dimension} & \multicolumn{2}{c|}{\emph{CUB200-2011}} & \multicolumn{2}{c|}{\emph{Stanford Dogs}} \\
  \cline{3-6} & & top1 & top5 & top1 & top5 \\
  \hline
  \hline
  VLAD ($k$=2) & 1,024 & 55.92 & 	62.51 & 	69.28 & 	74.43 \\
  VLAD ($k$=128) & 6,5536 & 55.66 & 	62.40 & 	68.47 & 	75.01 \\
  Fisher Vector ($k$=2) & 2,048 & 52.04 &	59.19 & 	68.37 & 	73.74 \\
  Fisher Vector ($k$=128) & 131,072 & 45.44 &	53.10 & 	61.40 & 	67.63 \\
  %SPoC & 256 & 43.41\% & 	50.97\% & 	56.89\% & 	63.87\%  \\
  avgPool & ~~512 & 56.42 & 	63.14 & 	73.76 & 	78.47  \\
  maxPool & ~~512 & 58.35 & 	64.18 & 	70.37 & 	75.59  \\
  \hline
  \hline
  avg\&maxPool & 1,024 & \textbf{59.72} & 	\textbf{65.79} & 	\textbf{74.86} & 	\textbf{79.24} \\
  \hline
 \end{tabular}
\end{table}

\subsection{Multiple Layer Ensemble}\label{sec:multilayer}
As studied in~\cite{Bharath15CVPR,Jonathan15CVPR}, the ensemble of multiple layers boosts the final performance. Thus, we also incorporate another SCDA feature produced from the $\text{relu}_{5\_2}$ layer which is three layers in front of $\text{pool}_5$ in the VGG-16 model~\cite{vgg16}.

\begin{figure}[t]
 \centering
 \subfloat[$M$ of $\rm pool_5$]  { \includegraphics[width=0.32\columnwidth]{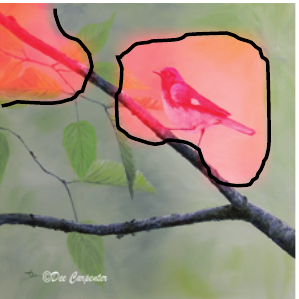} \label{fig:L28_1} }
 \qquad
 \subfloat[$\widetilde{M}$ of $\rm pool_5$] { \includegraphics[width=0.32\columnwidth]{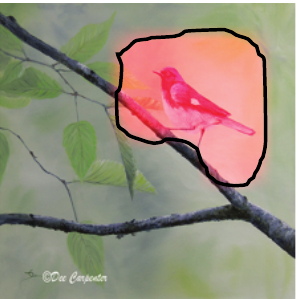} \label{fig:L28_2} }
 \qquad
 \subfloat[$M$ of $\rm relu_{5\_2}$] { \includegraphics[width=0.32\columnwidth]{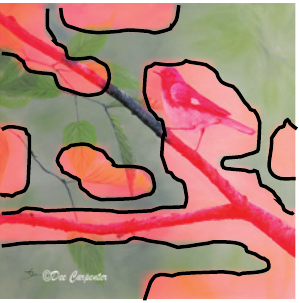} \label{fig:L28_3} }
 \qquad
 \subfloat[$\widetilde{M}_{\text{pool}_5} \cap M_{\text{relu}_{5\_2}}$] { \includegraphics[width=0.32\columnwidth]{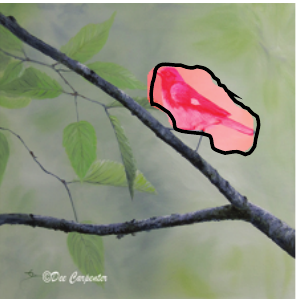} \label{fig:L28_4} }
  \caption{The mask map and its corresponding largest connected component of $\rm pool_5$, and the mask map and the final mask map (i.e., $\widetilde{M}_{\text{pool}_5} \cap M_{\text{relu}_{5\_2}}$) of $\rm relu_{5\_2}$. (The figure is best viewed in color.)}
 \label{fig:L28}
\end{figure}

Following $\text{pool}_5$, we get the mask map $M_{\text{relu}_{5\_2}}$ from $\text{relu}_{5\_2}$. Its activations are less related to the semantic meaning than those of $\text{pool}_5$. As shown in Fig.~\ref{fig:L28}~(c), there are many noisy parts. However, the bird is more accurately detected than $\text{pool}_5$. Therefore, we combine $\widetilde{M}_{\text{pool}_5}$ and $M_{\text{relu}_{5\_2}}$ together to get the final mask map of $\text{relu}_{5\_2}$. $\widetilde{M}_{\text{pool}_5}$ is firstly upsampled to the size of $M_{\text{relu}_{5\_2}}$. We keep the descriptors when their position in both $\widetilde{M}_{\text{pool}_5}$ and $M_{\text{relu}_{5\_2}}$ are 1, which are the final selected $\text{relu}_{5\_2}$ descriptors. The aggregation process remains the same. Finally, we concatenate the SCDA features of $\text{relu}_{5\_2}$ and $\text{pool}_5$ into a single representation, denoted by ``$\text{SCDA}^{+}$'':
\begin{equation}
\text{SCDA}^{+} \leftarrow \left[\text{SCDA}_{\text{pool}_5},~ \alpha \times \text{SCDA}_{\text{relu}_{5\_2}}\right]\,,
\end{equation}
where $\alpha$ is the coefficient for $\text{SCDA}_{\text{relu}_{5\_2}}$. It is set to 0.5 for FGIR. After that, we do the $\ell_2$ normalization on the concatenation feature. In addition, another $\text{SCDA}^{+}$ of the horizontal flip of the original image is incorporated, which is denoted as ``$\text{SCDA\_flip}^{+}$'' (4,096-d). Additionally, we also try to combine features from more different layers, e.g., pool$_4$. However, the retrieval performance improved slightly (about 0.01\%$\sim$0.04\% top-1 mAP), while the feature dimensionality became much larger than the proposed SCDA features.

\section{Experiments and Results}\label{sec:experiment}
In this section, we firstly describe the datasets and the implementation details of the experiments. Then, we report the fine-grained image retrieval results. We also test our proposed SCDA method on two general-purpose image retrieval datasets. As additional evidence to prove the effectiveness of SCDA, we report the fine-grained classification accuracy by fine-tuning the pre-trained model with image-level labels. Finally, the main observations are summarized.

\subsection{Datasets and Implementation Details}
For fine-grained image retrieval, the empirical evaluation is performed on six benchmark fine-grained datasets, \emph{CUB200-2011}~\cite{WahCUB200_2011} (200 classes, 11,788 images), \emph{Stanford Dogs}~\cite{Khosla11stanforddogs} (120 classes, 20,580 images), \emph{Oxford Flowers 102}~\cite{Flowers08} (102 classes, 8,189 images), \emph{Oxford-IIIT Pets}~\cite{Pets12} (37 classes, 7,349 images), \emph{Aircrafts}~\cite{airplanes} (100 classes, 10,000 images) and \emph{Cars}~\cite{cars} (196 classes, 16,185 images).

Additionally, two standard image retrieval datasets (\emph{INRIA Holiday}~\cite{Holiday08ECCV} and \emph{Oxford Building 5K}~\cite{Oxfordbuilding}) are employed for evaluating the general-purpose retrieval performance.

In experiments, for the pre-trained deep model, the publicly available VGG-16 model~\cite{vgg16} is employed to extract deep convolutional descriptors using the open-source library MatConvNet~\cite{matconvnet}. For all the retrieval datasets, the subtracted mean pixel values for zero-centering the input images are provided by the pre-trained VGG-16 model. All the experiments are run on a computer with Intel Xeon E5-2660 v3, 500G main memory, and an Nvidia Tesla K80 GPU.

\subsection{Fine-Grained Image Retrieval Performance}\label{sec:mainresults}

In the following, we report the results for fine-grained image retrieval. We compare the proposed method with several baseline approaches and three state-of-the-art general image retrieval approaches, \emph{SPoC}~\cite{Artem15ICCV}, \emph{CroW}~\cite{Yannis16arxiv} and \emph{R-MAC}~\cite{rmac}. The top-$1$ and top-5 mAP results are reported in Table~\ref{table:retrieval}.

Firstly, we conduct the SIFT descriptors with Fisher Vector encoding as the handcrafted-feature-based retrieval baseline. The parameters of SIFT and FV used in experiments followed~\cite{Lingxi15TMM}. The feature dimension is 32,768. Its retrieval performance on \textit{CUB200-2011}, \textit{Stanford Dogs}, \textit{Oxford Flowers} and \textit{Oxford Pets} is significantly worse than the deep learning methods/baselines. But, the retrieval results on rigid bodies like aircrafts and cars are good, while they are still worse than deep learning retrieval methods. In addition, we also feed the ground truth bounding boxes to replace the whole images. As shown in Table~\ref{table:retrieval}, because the ground truth bounding boxes of these fine-grained images just contain the main objects, ``SIFT\_FV\_gtBBox'' achieves significantly better performance than that of the whole images.

For the $\text{fc}_8$ baseline, because it requires the input images at a fixed size, the original images are resized to $224\times 224$ and then fed into VGG-16. Similar to the SIFT baseline, we also feed the ground truth bounding boxes to replace the whole images. The $\text{fc}_8$ feature of the ground truth bounding box achieves better performance. Moreover, the retrieval results of the $\text{fc}_8$ feature using the bounding boxes predicted by our method are also shown in Table~\ref{table:retrieval}, which are slightly lower than the ground-truth ones. This observation validates the effectiveness of our method's object localization once again.

For the $\text{pool}_5$ baseline, the $\text{pool}_5$ descriptors are extracted directly without any selection process. We pool them by both average- and max-pooling, and concatenate them into a 1,024-d representation. As shown in Table~\ref{table:retrieval}, the performance of $\text{pool}_5$ is better than ``$\text{fc}_8$\_im'', but much worse than the proposed SCDA feature. In addition, VLAD and FV is employed to encode the selected deep descriptors, and we denote the two methods as ``selectVLAD'' and ``selectFV'' in Table~\ref{table:retrieval}. The features of selectVLAD and selectFV have larger dimensionality, but lower mAP in the retrieval task.

State-of-the-art general image retrieval approaches, e.g., SPoC, CroW and R-MAC, can not get satisfactory results for fine-grained images. Hence, general deep learning image retrieval methods could not be directly applied to FGIR.

We also report the results of $\text{SCDA}^{+}$ and $\text{SCDA\_{flip}}^{+}$ on these six fine-grained datasets in Table~\ref{table:retrieval}. In general, $\text{SCDA\_{flip}}^{+}$ is the best amongst the compared methods. Comparing these results with the ones of SCDA, we find the multiple layer ensemble strategy (cf. Sec.~\ref{sec:multilayer}) could improve the retrieval performance, and furthermore horizontal flip boosts the performance significantly. Therefore, if your retrieval tasks prefer a low dimensional feature representation, SCDA is the optimal choice, or, the post-processing on $\text{SCDA\_{flip}}^{+}$ features is recommended.

\begin{table*}[th!]
 \caption{Comparison of fine-grained image retrieval performance. The best result of each column is in bold.} \label{table:retrieval}
 \centering
 \begin{tabular}{|c|c||c|c||c|c||c|c||c|c||c|c||c|c|}
  \hline
  \multirow{2}{*}{Method} & \multirow{2}{*}{Dimension} & \multicolumn{2}{c||}{\emph{CUB200-2011}} & \multicolumn{2}{c||}{\emph{Stanford Dogs}} & \multicolumn{2}{c||}{\emph{Oxford Flowers}} & \multicolumn{2}{c||}{\emph{Oxford Pets}} & \multicolumn{2}{c||}{\emph{Aircrafts}} & \multicolumn{2}{c|}{\emph{Cars}}\\
  \cline{3-14} & & top1 & top5 & top1 & top5 & top1 & top5 & top1 & top5 & top1 & top5 & top1 & top5\\
  \hline
  \hline
  SIFT\_FV & 32,768 &	5.25 & 	8.07 & 	12.58 & 	16.38 & 	30.02 & 	36.19 & 	17.50 & 	24.97 & 30.69 &	37.44 &	19.30 &	24.11  \\
  SIFT\_FV\_gtBBox & 32,768 &	9.98 & 	14.29 & 	15.86 & 	21.15 & 	-- & 	-- & 	-- & 	-- & 38.70 &	46.87 &	34.47 &	40.34  \\
  \hline
  $\text{fc}_8$\_im & 4,096 &	39.90 & 	48.10 & 	66.51 & 	72.69 & 	55.37 & 	60.37 & 	82.26 & 	86.02 & 28.98 &	35.00 &	19.52 &	25.77  \\
  $\text{fc}_8$\_gtBBox & 4,096&	47.55 & 	55.34 & 	70.41 & 	76.61 & --& --& --& -- & 34.80& 	41.25& 	30.02& 	37.45 \\ 
  $\text{fc}_8$\_predBBox & 4,096	&45.24 & 	53.05 & 	68.78 & 	74.09  & 	57.16 & 	62.24 & 	85.55 & 	88.47 & 30.42& 	36.50& 	22.27& 	29.24  \\ 
\hline
  $\text{pool}_5$ & 1,024 &	57.54 & 	63.66 &  69.98 &  75.55 &  			70.73 & 	74.05 & 	85.09 & 	87.74 & 47.37& 	53.61& 	34.88& 	41.86  \\
 % $\text{Pool}_5$\_predBBox & 512&	56.27\% & 	62.50\% & 	68.40\% & 	73.67\% & 	69.82\% & 	73.10\% & 	83.78\% & 	86.59\% \\
\hline
  selectFV & 2,048&	52.04 & 	59.19 & 	68.37 & 	73.74 & 	70.47 & 	73.60 & 	85.04 & 	87.09 & 48.69& 	54.68& 	35.32& 	41.60  \\
  selectVLAD & 1,024&	55.92 & 	62.51 & 	69.28 & 	74.43 & 	73.62 &	76.86 & 	85.50 & 	87.94 & 50.35& 	56.37& 	37.16& 	43.84 \\ 
\hline
  SPoC (w/o cen.) & 256 &	34.79 & 	42.54 & 	48.80 & 	55.95 &	71.36 & 	74.55 & 	60.86 & 	67.78 & 37.47& 	43.73& 	29.86& 	36.23  \\ 
  SPoC (with cen.) & 256 & 39.61 & 	47.30 & 	48.39 & 	55.69 &	65.86 & 	70.05 & 	64.05 & 	71.22 & 42.81& 	48.95& 	27.61& 	33.88  \\
  CroW & 256 &  53.45 & 	59.69 & 	62.18 & 	68.33 &	 73.67 & 	76.16 & 	76.34 & 	80.10 & 53.17& 	58.62& 	44.92& 	51.18  \\
  R-MAC&	512&	52.24 &	59.02& 	59.65& 	66.28& 	76.08& 	78.19& 	76.97& 	81.16& 	48.15& 	54.94 &	\textbf{46.54} &	\textbf{52.98} \\
\hline
\hline
  SCDA & 1,024&	59.72 & 	65.79 & 	74.86 & 	79.24 & 	75.13 & 	77.70 & 	87.63 & 	89.26 & 53.26& 	58.64 &	38.24 &	45.16  \\ 
  $\text{SCDA}^{+}$ & 2,048&	59.68 & 	65.83 & 	74.15 & 	78.54 & 	75.98 & 	78.49 & 	87.99 & 	89.49 & 53.53& 	59.11& 	38.70 &	45.65   \\
  $\text{SCDA\_flip}^{+}$ & 4,096&	\textbf{60.65} & 	\textbf{66.75} & 	\textbf{74.95} & 	\textbf{79.27} & 	\textbf{77.56} & 	\textbf{79.77} & 	\textbf{88.19} & 	\textbf{89.65} & \textbf{54.52}& 	\textbf{59.90}& 	40.12& 	46.73  \\ 
  \hline
 \end{tabular}
\end{table*}

\begin{table*}[th!]
 \caption{Comparison of different compression methods on ``$\text{SCDA\_{flip}}^{+}$''.} \label{table:retrievalsvd}
 \centering
 \begin{tabular}{|c|c||c|c||c|c||c|c||c|c||c|c||c|c|}
  \hline
  \multirow{2}{*}{Method} & \multirow{2}{*}{Dimension} & \multicolumn{2}{c||}{\emph{CUB200-2011}} & \multicolumn{2}{c||}{\emph{Stanford Dogs}} & \multicolumn{2}{c||}{\emph{Oxford Flowers}} & \multicolumn{2}{c||}{\emph{Oxford Pets}} & \multicolumn{2}{c||}{\emph{Aircrafts}} & \multicolumn{2}{c|}{\emph{Cars}}\\
  \cline{3-14} & & top1 & top5 & top1 & top5 & top1 & top5 & top1 & top5 & top1 & top5 & top1 & top5\\
  \hline
  \hline
  \multirow{2}{*}{PCA} & 256&	60.48 & 	66.55 & 	{74.63} & 	{79.09} & 	76.38 & 	79.32 & 	{87.82} & 	{89.75}  & 52.75& 	58.24& 	37.94& 	44.54   \\
  & 512 &	60.37 & 	66.78 & 	74.76 & 	79.27 & 	77.15 & 	79.50 & 	87.46 & 	89.71& 54.13& 	59.36& 	39.26& 	45.85    \\
  \hline
  \multirow{2}{*}{SVD} & 256 & 60.34 & 	66.57 & 	\textbf{74.79} & 	\textbf{79.27} & 	76.79 & 	79.32 & 	\textbf{87.84} & 	\textbf{89.79} & 52.90 	&58.20& 	38.04 &	44.57   \\
  & 512 & 60.41 & 	66.82 & 	74.72 & 	79.26 & 	77.10 & 	79.48 & 	87.41 & 	89.72 & 54.13 	&59.38 &	39.36 &	45.91    \\
  \hline
  \multirow{2}{*}{SVD$+$whitening} & 256 & \textbf{62.29} & 	\textbf{68.16} & 	71.57 & 	76.68 & 	80.74 & 	82.42 & 	85.47 & 	87.99 & 59.02& 	64.85& 	50.14 &	56.39 \\
  & 512 & {62.13} & 	68.13 & 	71.07 & 	76.06 & 	\textbf{81.44} & 	\textbf{82.82} & 	85.23 & 	87.62  & \textbf{61.21}& \textbf{66.49} &	\textbf{53.30} &	\textbf{59.11}  \\
  \hline
 \end{tabular}
\end{table*}

\subsubsection{Post-Processing} In the following, we compare several feature compression methods on the $\text{SCDA\_{flip}}^{+}$ feature: (a) Singular Value Decomposition (SVD); (b) Principal Component Analysis (PCA); (c) PCA whitening (its results were much worse than other methods and are omitted) and (d) SVD whitening. We compress the $\text{SCDA\_{flip}}^{+}$ feature to 256-d and 512-d, respectively, and report the compressed results in Table~\ref{table:retrievalsvd}. Comparing the results shown in Table~\ref{table:retrieval} and Table~\ref{table:retrievalsvd}, the compressed methods can reduce the dimensionality without hurting the retrieval performance. SVD (which does not remove the mean vector) has slightly higher rates than PCA (which removes the mean vector). The ``512-d SVD+whitening'' feature can achieve better retrieval performance: 2\%$\sim$4\% higher than the original $\text{SCDA\_{flip}}^{+}$ feature on \emph{CUB200-2011} and \emph{Oxford Flowers}, and significantly 7\%$\sim$13\% on \emph{Aircrafts} and \emph{Cars}. Moreover, ``512-d SVD+whitening'' with less dimensions generally achieves better performance than other compressed SCDA features. Therefore, we take it as our optimal choice for FGIR. In the following, we present some retrieval examples based on ``512-d SVD+whitening''.

\begin{figure*}[t]
 \centering
 \includegraphics[width=0.9\textwidth,height=40em]{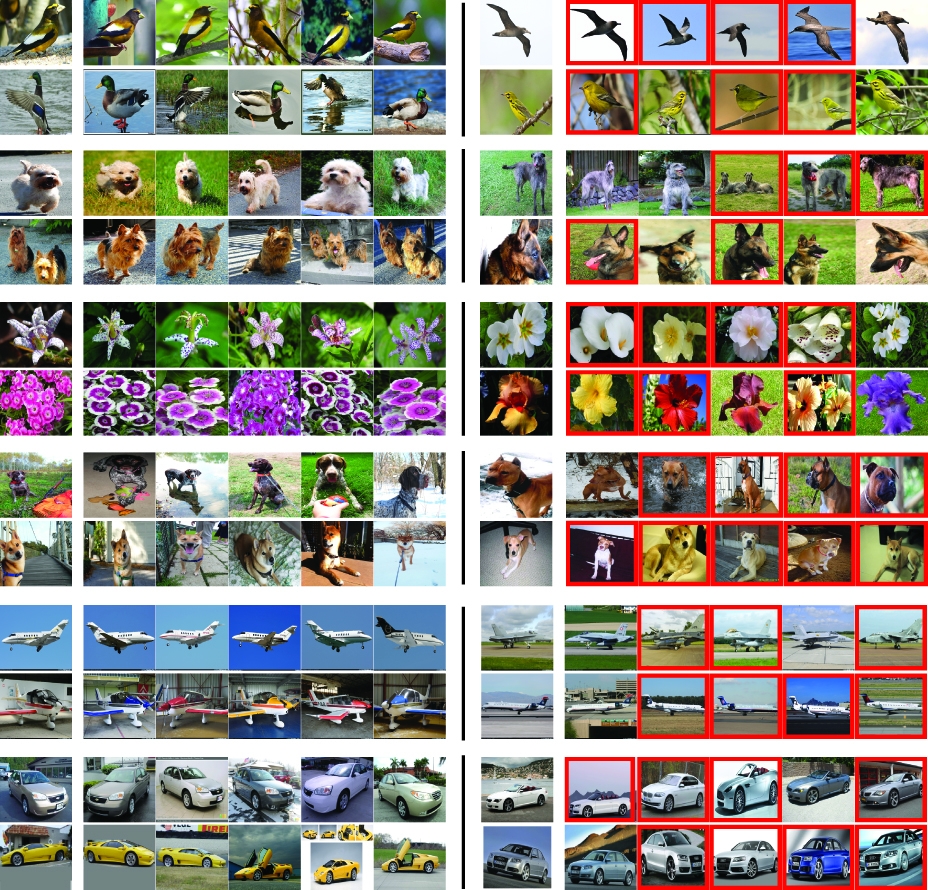}
  \caption{Some retrieval results of six fine-grained datasets. On the left, there are two successful cases for each datasets; while on the right, there are failure cases. The first image in each row is the query image. Wrong retrieval results are marked by red boxes. (The figure is best viewed in color.)}
 \label{fig:query}
\end{figure*}

In Fig.~\ref{fig:query}, we show two successful retrieval results and two failure cases for each fine-grained dataset, respectively. As shown in the successful cases, our method can work well when the same kind of birds, animals, flowers, aircrafts or cars appear in different kinds of background. In addition, for these failure cases, there exist only tiny differences between the query image and the returned ones, which can not be accurately detected in this pure unsupervised setting. We can also find some interesting observations, e.g., the last failure case of the flowers and pets. For the flowers, there are two correct predictions in the top-5 returned images. Even though the flowers in the correct predictions have different colors with the query, our method can still find them. For the pets' failure cases, the dogs in the returned images have the same pose as the query image.

\subsection{Quality and Insight of the SCDA Feature}
\begin{figure*}[t]
 \centering
 \includegraphics[width=0.9\textwidth,height=15em]{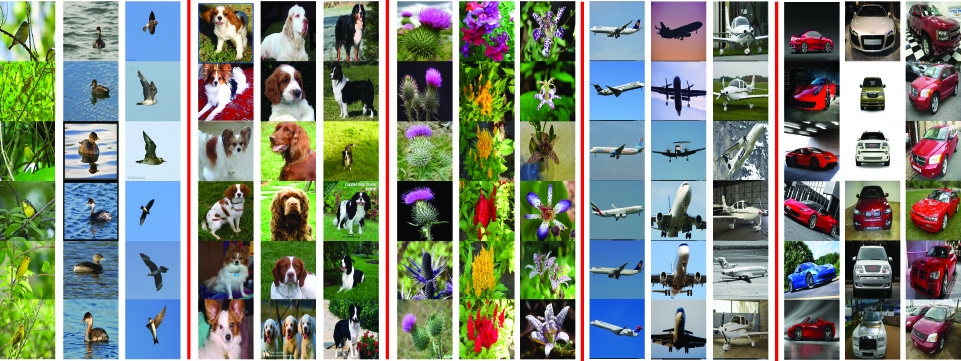}
  \caption{Quality demonstrations of the SCDA feature. From the top to bottom of each column, there are six returned original images in the descending order of one sorted dimension of ``256-d SVD+whitening''. (Best viewed in color and zoomed in.)}
 \label{fig:svd}
\end{figure*}

In this section, we discuss the quality of the proposed SCDA feature. After SVD and whitening, the former distributed dimensions of SCDA have more discriminative ability, i.e., directly correspond to semantic visual properties that are useful for retrieval. We use five datasets (\emph{CUB200-2011}, \emph{Stanford Dogs}, \emph{Oxford Flowers}, \emph{Aircrafts} and \emph{Cars}) as examples to illustrate the quality. We first select one dimension of  ``512-d SVD+whitening'', and then sort the value of that dimension in the descending order. Then, we visualize images in the same order, which is shown in Fig.~\ref{fig:svd}.

Images of each column have some similar ``attributes'', e.g., \emph{living in water} and \emph{opening wings} for birds; \emph{brown and white heads} and \emph{similar looking faces} for dogs; \emph{similar shaped inflorescence} and \emph{petals with tiny spots} for flowers; \emph{similar poses} and \emph{propellers} for aircrafts; \emph{similar point of views} and \emph{motorcycle types} for cars. Obviously, the SCDA feature has the ability to describe the main objects' attributes (even subtle attributes). Thus, it can produce human-understandable interpretation manuals for fine-grained images, which might explain its success in fine-grained image retrieval. In addition, because the values of the compressed SCDA features might be positive, negative and zero, it is meaningful to sort these values either in the descending order (shown in Fig.~\ref{fig:svd}), or ascending order. The images returned in ascending also exhibit some similar visual attributes.

\subsection{General Image Retrieval Results}
For further investigation of the effectiveness of the proposed SCDA method, we compare it with three state-of-the-art general image retrieval approaches ({SPoC}~\cite{Artem15ICCV}, {CroW}~\cite{Yannis16arxiv} and {R-MAC}~\cite{rmac}) on the \emph{INRIA Holiday}~\cite{Holiday08ECCV} and \emph{Oxford Building 5K}~\cite{Oxfordbuilding} datasets. Following the protocol in~\cite{Artem14ECCV,Artem15ICCV}, for the \emph{Holiday} dataset, we manually fix images in the wrong orientation by rotating them by $\pm90$ degrees, and report the mean average precision (mAP) over 500 and 55 queries for ~\emph{Holiday} and \emph{Oxford Building 5K}, respectively.

In the experiments of these two general image retrieval datasets, we use the SCDA and SCDA\_flip features, and compress them by SVD whitening. As shown by the results presented in Table~\ref{table:retrievalgnr}, the compressed SCDA\_flip (512-d) achieves the highest mAP among the proposed ones. In addition, comparing with state-of-the-arts, the compressed SCDA\_flip (512-d) is significantly better than {SPoC}~\cite{Artem15ICCV}, {CroW}~\cite{Yannis16arxiv}, and comparable with the R-MAC approach~\cite{rmac}. Therefore, the proposed SCDA not only significantly outperforms the general image retrieval state-of-the-art approaches for fine-grained image retrieval, but can also obtain comparable results for general-purpose image retrieval tasks.

\begin{table}[t]
 \caption{Comparison of general image retrieval performance. The best result of each column is marked in bold.} \label{table:retrievalgnr}
 \centering
 \begin{tabular}{|c|r|c|c|}
  \hline
  Method & Dim. & \emph{Holiday} & \emph{Oxford Building} \\
  \hline
  \hline
  SPoC (w/o cen.)~\cite{Artem15ICCV} & 256 & 80.2&	58.9\\
  SPoC (with cen.)~\cite{Artem15ICCV} &256 & 78.4&	65.7\\
  CroW~\cite{Yannis16arxiv} & 256 & 83.1&	65.4 \\
  R-MAC~\cite{rmac} & 512 & \textbf{92.6} & 66.9  \\
  Method of~\cite{Zheng16arxiv} & 9,664 & 84.2 & \textbf{71.3}\\
  \hline
  \hline
  SCDA & 1,024 & 90.2	&61.7 \\
  SCDA\_flip & 2,048 & 90.6 & 	62.5 \\
  SCDA\_flip (SVD whitening) & 256 & 91.6 & 	66.4\\
  SCDA\_flip (SVD whitening)& 512 & {92.1} & 	{67.7} \\
  \hline
 \end{tabular}
\end{table}

\subsection{Fine-Grained Classification Results}
In the end, we compare with several state-of-the-art fine-grained \emph{classification} algorithms to validate the effectiveness of SCDA from the classification perspective.

In the classification experiments, we adopt two strategies to fine-tune the VGG-16 model with only the image-level labels. One strategy is directly fine-tuning the pre-trained model of the original VGG-16 architecture by adding the horizontal flips of the original images as data augmentation. After obtaining the fine-tuned model, we extract the $\text{SCDA\_{flip}}^{+}$ as the whole image representations and feed them into a linear SVM~\cite{REF08a} to train a classifier. 

The other strategy is to build an end-to-end SCDA architecture. Before each epoch, the masks $\widetilde{M}$ of $\rm pool_5$ and $\rm relu_{5\_2}$ are extracted first. Then, we implement the selection process as an element-wise product operation between the convolutional activation tensor $T$ and the mask matrix $\widetilde{M}$. Therefore, the descriptors located in the object region will remain,  while the other descriptors will become zero vectors. In the forward pass of the end-to-end SCDA, we select the descriptors of $\rm pool_5$ and $\rm relu_{5\_2}$ as aforementioned, and then, both max- and average-pool (followed by $\ell_2$ normalization) the selected descriptors into the corresponding SCDA feature. After that, the SCDA features of $\rm pool_5$ and $\rm relu_{5\_2}$ are concatenated, which is the so called ``SCDA$^+$'', as the final representation of the end-to-end SCDA model. Then, a classification (fc+softmax) layer is added for end-to-end training. Because the partial derivative of the mask is zero, it will not affect the backward  processing of the end-to-end SCDA. After each epoch, the masks will be updated based on the learned SCDA model in the last epoch. When end-to-end SCDA converges, the $\text{SCDA\_{flip}}^{+}$ is also extracted.

For both strategies, the coefficient $\alpha$ of $\text{SCDA\_{flip}}^{+}$ is set to 1 to let the classifier to learn and then select important dimensions automatically. The classification accuracy comparison is listed in Table~\ref{table:acc}. 

For the first fine-tuning strategy, the classification accuracy of our method (``SCDA (f.t.)'') is comparable or even better than the algorithms trained with strong supervised annotations, e.g.,~\cite{Ning14ECCV,Di15CVPR}. For these algorithms using only image-level labels, our classification accuracy is comparable with the algorithms using similar fine-tuning strategies (\cite{Tianjun15CVPR,Marcel15ICCV,TIP2016weakly}), but still does not perform as well as those using more powerful deep architectures and more complicated data augmentations~\cite{Tsung-Yu15ICCV,Max15NIPS}. For the second fine-tuning strategy, even though ``SCDA (end-to-end)'' obtains a slightly lower classification accuracy than ``SCDA (f.t.)'', the end-to-end SCDA model contains the least number of parameters (i.e., only 15.53M), which attributes to no fully connected layers in the architecture.

\begin{table*}[t]
 \caption{Comparison of classification accuracy on six fine-grained datasets. The ``SCDA (f.t.)'' denotes the SCDA features are extracted from the directly fine-tuned VGG-16 model. The ``SCDA (end-to-end)'' represents the SCDA features are from the fine-tuned end-to-end SCDA model.} \label{table:acc}
 \centering
 \setlength{\tabcolsep}{3.5pt}
 \begin{tabular}{|c|c|c|c|c|c|r|r||c|c|c|c|c|c|}
  \hline
 \multirow{2}{*}{Method} & \multicolumn{2}{c|}{Train phase} & \multicolumn{2}{c|}{Test phase} & \multirow{2}{*}{Model} & \multirow{2}{*}{$\sharp$ Para.~} & \multirow{2}{*}{Dim.~} & \multirow{2}{*}{\emph{Birds}}  &  \multirow{2}{*}{\emph{Dogs}} &  \multirow{2}{*}{\emph{Flowers}} &  \multirow{2}{*}{\emph{Pets}} &  \multirow{2}{*}{\emph{Aircrafts}} &  \multirow{2}{*}{\emph{Cars}} \\
  \cline{2-5} & BBox & Parts & BBox & Parts & & & & & & & & &\\
  \hline
  \hline
  PB R-CNN with BBox~\cite{Ning14ECCV} & $\checkmark$ & $\checkmark$ & $\checkmark$ & & Alex-Net$\times$3 & 173.03M & 12,288 & 76.4 &-- &-- &-- &-- &--\\
  Deep LAC~\cite{Di15CVPR} & $\checkmark$ & $\checkmark$ & $\checkmark$ & & Alex-Net$\times 3$ & 173.03M & 12,288 & 80.3&-- &-- &--&-- &--  \\
  PB R-CNN~\cite{Ning14ECCV} & $\checkmark$ & $\checkmark$ & & & Alex-Net$\times 3$ & 173.03M &12,288 & 73.9&-- &-- &--&-- &-- \\  
  Two-Level~\cite{Tianjun15CVPR} & & & & &VGG-16$\times 1$ & 135.07M & 16,384  & 77.9 & -- &-- &--&-- &--  \\
  Weakly supervised FG~\cite{TIP2016weakly} & & & & &VGG-16$\times 1$ & 135.07M & 262,144  & 79.3 & 80.4 &-- &--&-- &--  \\
  Constellations~\cite{Marcel15ICCV} & & & & &VGG-19$\times 1$ & 140.38M & 208,896 & 81.0 & 68.6\footnotemark[1] & 95.3 & 91.6&-- &-- \\
  Bilinear~\cite{Tsung-Yu15ICCV} & & & & &VGG-16 and VGG-M & 73.67M & 262,144 & 84.0&-- &-- &-- & 83.9 & 91.3 \\
  Spatial Transformer Net~\cite{Max15NIPS} & & & & & ST-CNN (inception)$\times 4$ & 62.68M & {4,096} & 84.1 &-- &-- &--&-- &--\\
  \hline
  \hline
  SCDA (f.t.) & & & & & VGG-16$\times$1 & 135.07M & {4,096} & {80.5} & {78.7} & {92.1} & {91.0} & {79.5} & {85.9} \\
  SCDA (end-to-end) & & & & & VGG-16 (w/o FCs)$\times$1& 15.53M & {4,096} & {80.1} & {77.4} & {90.2} & {90.3} & {78.6} & {85.1} \\
  \hline
 \end{tabular}
\footnotemark[1]{\scriptsize{\cite{Marcel15ICCV} reported the result of the \emph{Birds} dataset using VGG-19, while the result of \emph{Dogs} is based on the Alex-Net model.}}
\end{table*}

Thus, our method has less dimensions and is simple to implement, which makes SCDA more scalable for large-scale datasets without strong annotations and is easier to generalize. In addition, the CroW~\cite{Yannis16arxiv} paper presented the classification accuracy on \emph{CUB200-2011} without any fine-tuning (56.5\% by VGG-16). We also experiment on the 512-d SCDA feature (only contains the max-pooling part this time for fair comparison) without any fine-tuning. The classification accuracy on that dataset is 73.7\%, which outperforms their performance by a large margin.

\subsection{Additional Experiments on Completely Disjoint Classes}
For further investigating the generalization ability of the proposed SCDA method, we additionally conduct experiments on a recently released fine-grained dataset for biodiversity analysis, i.e., the \emph{Moth} dataset~\cite{Moth15CVPR}. This dataset includes 2,120 moth images of 675 highly similar classes, which are completely disjoint with the images of ImageNet. In Table~\ref{table:retrievalmoth}, we present the retrieval results of SCDA and other baseline methods. Because there are several classes in \emph{Moth} have less than five images per class, we only report the top-1 mAP results. Consistent with the observations in Sec.~\ref{sec:mainresults}, $\text{SCDA\_flip}^{+}$ still outperforms other baseline methods, which proves the proposed method could generalize well.

\begin{table}[t!]
 \caption{Comparison of retrieval performance on the \emph{Moth} dataset~\cite{Moth15CVPR}. The best result is marked in bold. Note that, because there are several fine-grained categories of \emph{Moth} containing less than five images for each category, we here only report the top-1 mAP results.} \label{table:retrievalmoth}
 \centering
 \begin{tabular}{|c|c||c|}
  \hline
  {Method} & {Dimension} & Top-1 mAP \\
  \hline
  \hline
  $\text{fc}_8$\_im & 4,096 &	42.52   \\
\hline
  $\text{pool}_5$ & 1,024 &	42.67  \\
\hline
  selectFV & 2,048 &	40.33   \\
  selectVLAD & 1,024 &	42.41    \\
\hline
  %SPoC (w/o cen.) & 256 &	   \\ 
  SPoC (with cen.) & 256 & 42.96   \\
  CroW & 256 &  50.78   \\
  R-MAC &	512& 45.38 \\
\hline
\hline
  SCDA & 1,024 &	47.48  \\ 
  $\text{SCDA}^{+}$ & 2,048 &	49.78    \\
  $\text{SCDA\_flip}^{+}$ & 4,096 &	{50.52}   \\ 
\hline
\hline
  $\text{SCDA\_flip}^{+}$ (SVD whitening) & 256 &	{54.96}   \\ 
  $\text{SCDA\_flip}^{+}$ (SVD whitening) & 512 &	\textbf{57.19}   \\
\hline
 \end{tabular}
\end{table}

\subsection{Computational Time Comparisons}

In this section, we compare the inference speeds of our SCDA with other methods. Because the methods listed in Table~\ref{table:infertime} can handle arbitrary image resolutions, different fine-grained data sets have different speeds. Specifically, much larger image will cause the GPUs out of memory. Thus, according to the original image scaling, we resize the images until $\rm \min(im\_height,im\_width)=700$ pixels, when $\rm \min(im\_height,im\_width)>700$. As the speeds reported in Table~\ref{table:infertime}, it is understandable that the speed of SCDA is lower than that of pool$_5$. In general, SCDA has the comparable computational speeds with CroW, and is significantly faster than R-MAC. But, its speed is slightly lower (about 1 frame/sec) than SPoC. In practice, if your retrieval tasks prefer high accuracy, SCDA\_flip$^+$ is recommended. While, if you prefer efficiency, SCDA is scalable enough for handling large scale fine-grained datasets. Meanwhile, SCDA will bring good retrieval accuracy (cf. Table~\ref{table:retrieval}).

\begin{table}[th!]
 \caption{Comparisons of inference speeds (frames/sec) on six fine-grained image datasets.} \label{table:infertime}
 \centering
 \setlength{\tabcolsep}{3.5pt}
 \begin{tabular}{|c|c|c|c|c|c|c|}
  \hline
  {Method}  & {\emph{Birds}} & {\emph{Dogs}} & {\emph{Flowers}} & {\emph{Pets}} & {\emph{Aircrafts}} & {\emph{Cars}}\\  
  \hline
  \hline
  pool$_5$ & 9.54 & 	9.01	 & 6.15	 & 10.31 & 	2.92	 & 5.81 \\
  SPoC (w/o cen.) &  8.70 & 	8.77 & 	5.92	 & 10.10 & 	2.76 & 	5.46 \\
  SPoC (with cen.)  & 8.40	 & 8.62	 & 5.78	 & 10.10	 & 2.72	 & 5.49 \\
  CroW  & 7.81 & 	7.04	 & 5.26 & 	7.75	 & 2.60	 & 4.72 \\
  R-MAC  & 4.22 & 	4.52 & 	3.00	 & 5.05	& 1.93	& 3.62 \\
  \hline
  \hline
  SCDA & 9.09  &  7.81  &  4.85  &  9.61  &  2.05  &  4.16 \\
  $\text{SCDA}^{+}$  &  7.46  &   6.66 &    3.34   &  7.14   &  1.11  &   2.35 \\
  $\text{SCDA\_flip}^{+}$ &  3.80   &  3.48&    1.81  &   3.83  &   0.55  &   1.19 \\
  \hline
 \end{tabular}
\end{table}

\subsection{Summary of Experimental Results}

In the following, we summarize several empirical observations of the proposed selective convolutional descriptor aggregation method for FGIR.
\begin{itemize}
\item Simple aggregation methods such as max- and average-pooling achieved better retrieval performance than high-dimensional encoding approaches. The proposed SCDA representation concatenated both the max- and average-pooled features, which achieved the best retrieval performance as reported in Table~\ref{table:poolcompa} and Table~\ref{table:retrieval}.
\item Convolutional descriptors performed better than the representations of the fully connected layer for FGIR. In Table~\ref{table:retrieval}, the representations of ``$\rm pool_5$'', ``selectFV'' and ``selectVLAD'' are all based on the convolutional descriptors. No matter what kind of aggregation methods they used, their top-$k$ retrieval results are (significantly) better than the fully connected features.
\item Selecting descriptors is beneficial to both fine-grained image retrieval and general-purposed image retrieval. As the results reported in Table~\ref{table:retrieval} and Table~\ref{table:retrievalgnr}, the proposed SCDA method achieved the best results for FGIR, meanwhile was comparable with general image retrieval state-of-the-art approaches.
\item The SVD whitening compression method can not only reduce the dimensions of the SCDA feature, but also improve the retrieval performance, even by a large margin (cf. the results of \emph{Aircrafts} and \emph{Cars} in Table~\ref{table:retrievalsvd}). Moreover, the compressed SCDA feature had the ability to describe the main objects' subtle attributes, which is shown in Fig.~\ref{fig:svd}.
\end{itemize}

\section{Conclusions}\label{sec:concl}
In this paper, we proposed to solely use a CNN model pre-trained on non-fine-grained tasks to tackle the novel and difficult fine-grained image retrieval task. We proposed the Selective Convolutional Descriptor Aggregation (SCDA) method, which is unsupervised and does not require additional learning. SCDA first localized the main object in fine-grained image unsupervised with high accuracy. The selected (localized) deep descriptors were then aggregated using the best practices we found to produce a short feature vector for a fine-grained image. Experimental results showed that, for fine-grained image retrieval, SCDA outperformed all the baseline methods including general image retrieval state-of-the-arts. Moreover, these features of SCDA exhibited well-defined semantic visual attributes, which may explain why it has high retrieval accuracy for fine-grained images. Meanwhile, SCDA had the comparable retrieval performance on standard general image retrieval datasets. The satisfactory results of both fine-grained and general-purpose image retrieval datasets validated the benefits of selecting convolutional descriptors.

In the future, we consider including the selected deep descriptors' weights to find object parts. Another interesting direction is to explore the possibility of pre-trained models for more complicated vision tasks such as unsupervised object segmentation. Indeed, enabling models trained for one task to be \textit{reusable} for another different task, particularly without additional training, is an important step toward the development of \emph{learnware}~\cite{Zhou2016Leanware}. 

% use section* for acknowledgment
%\section*{Acknowledgment}

%The authors would like to thank the anonymous reviewers, whose comments have helped improving this paper. 

\bibliographystyle{IEEEtran}
\bibliography{IEEEabrv,SCDA}

% biography section

\begin{IEEEbiography}[{\includegraphics[width=1in,height=1.25in,clip,keepaspectratio]{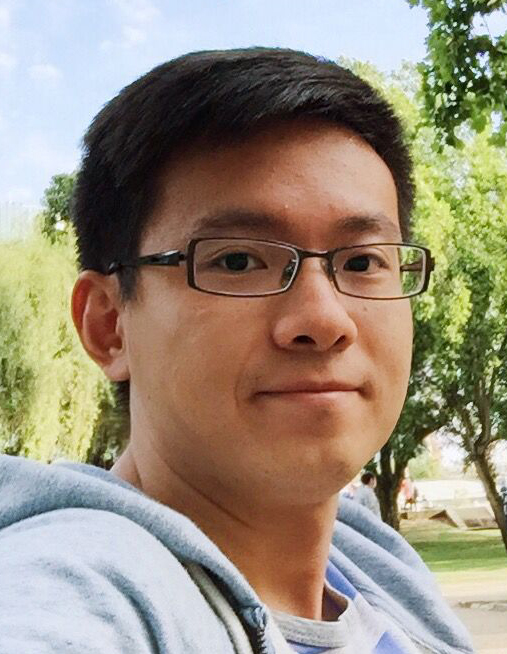}}]{Xiu-Shen Wei}
received the BS degree in Computer Science and Technology in 2012. He is currently a PhD candidate in the Department of Computer Science and Technology at Nanjing University, China. He achieved the first place in the Apparent Personality Analysis competition (in association with ECCV 2016) and the first runner-up in the Cultural Event Recognition competition (in association with ICCV 2015) as the team director. His research interests are computer vision and machine learning.
\end{IEEEbiography}

\begin{IEEEbiography}[{\includegraphics[width=1in,height=1.25in,clip,keepaspectratio]{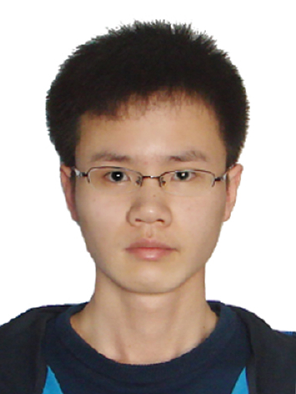}}]{Jian-Hao Luo}
received his BS degree in the College of Computer Science and Technology from Jilin University, China, in 2015. He is currently working toward the PhD degree in the Department of Computer Science and Technology, Nanjing University, China. His research interests are computer vision and machine learning.
\end{IEEEbiography}

\begin{IEEEbiography}[{\includegraphics[width=1in,height=1.25in,clip,keepaspectratio]{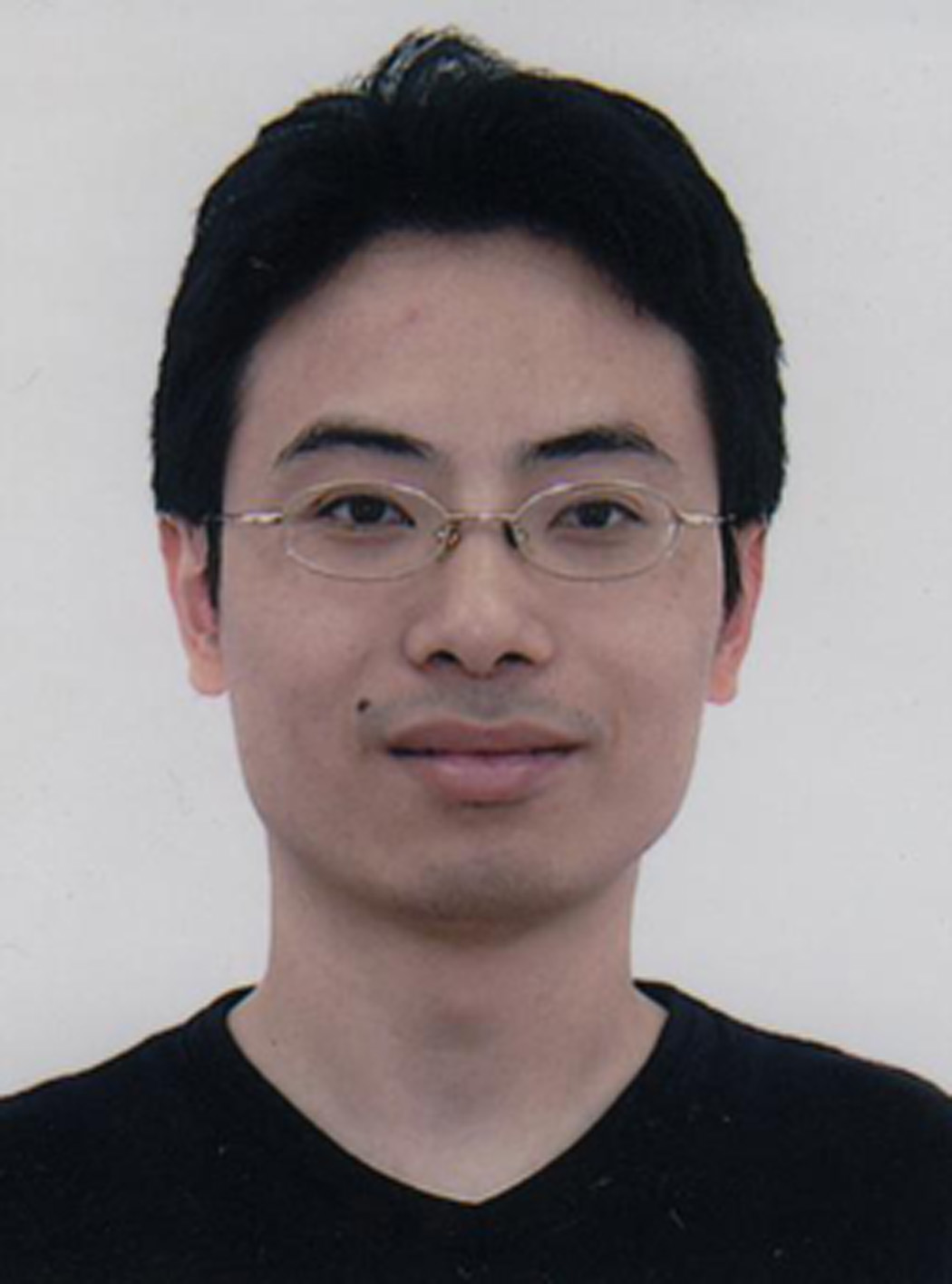}}]{Jianxin Wu}(M'09)
received his BS and MS degrees in computer science from Nanjing University, and his PhD degree in computer science from the Georgia Institute of Technology. He is currently a professor in the Department of Computer Science and Technology at Nanjing University, China, and is associated with the National Key Laboratory for Novel Software Technology, China. He was an assistant professor in the Nanyang Technological University, Singapore, and has served as an area chair for CVPR 2017, ICCV 2015, senior PC member for AAAI 2017, AAAI 2016 and an associate editor for Pattern Recognition Journal. His research interests are computer vision and machine learning. He is a member of the IEEE.
\end{IEEEbiography}

\begin{IEEEbiography}[{\includegraphics[width=1in,height=1.25in,clip,keepaspectratio]{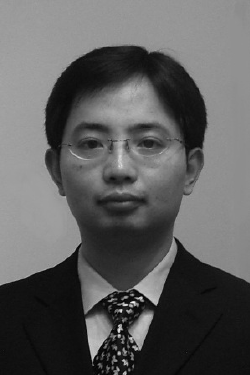}}]{Zhi-Hua Zhou} (S'00-M'01-SM'06-F'13) received the BSc, MSc and PhD degrees in computer science from Nanjing University, China, in 1996, 1998 and 2000, respectively, all with the highest honors. He joined the Department of Computer Science \& Technology at Nanjing University as an Assistant Professor in 2001, and is currently Chair Professor and Standing Deputy Director of the National Key Laboratory for Novel Software Technology; he is also the Founding Director of the LAMDA group. His research interests are mainly in artificial intelligence, machine learning, data mining and pattern recognition. He has authored the book ``Ensemble Methods: Foundations and Algorithms'', and published more than 100 papers in top-tier international journals or conference proceedings. He has received various awards/honors including the National Natural Science Award of China, the IEEE CIS Outstanding Early Career Award, the Microsoft Professorship Award, the Fok Ying Tung Young Professorship Award, and twelve international journals/conferences papers or competitions awards. He also holds 15 patents. He is an Executive Editor-in-Chief of the \textit{Frontiers of Computer Science}, Associate Editor-in-Chief of the \textit{Science China: Information Sciences}, Associate Editor of the \textit{ACM Transactions on Intelligent Systems and Technology}, \textit{IEEE Transactions on Neural Networks and Learning Systems}, etc. He served as Associate Editor-in-Chief for \textit{Chinese Science Bulletin} (2008-2014), Associate Editor for \textit{IEEE Transactions on Knowledge and Data Engineering} (2008-2012) and \textit{Knowledge and Information Systems} (2003-2008). He founded ACML (Asian Conference on Machine Learning), served as Advisory Committee member for IJCAI, Steering Committee member for PAKDD and PRICAI, and Chair of various conferences such as General co-chair of ACML 2012, PCM 2013, PAKDD 2014, Program co-chair of SDM 2013, ICDM 2015, IJCAI 2015 Machine Learning Track and AAAI 2019, Workshop co-chair of KDD 2012, ICDM 2014, Tutorial co-chair of CIKM 2014, KDD 2015, and Area chair of ICML, NIPS, etc. He is the Chair of the IEEE CIS Data Mining and Big Data Analytics Technical Committee, the Chair of the CAAI (Chinese Association of Artificial Intelligence) Machine Learning Society, the Chair of the CCF (China Computer Federation) Artificial Intelligence \& Pattern Recognition Society, and the Chair of the IEEE Computer Society Nanjing Chapter.  He is a Fellow of the ACM, AAAS, AAAI, IEEE, IAPR, IET/IEE and CCF.
\end{IEEEbiography}

\end{document}